# A large comparison of feature-based approaches for buried target classification in forward-looking ground-penetrating radar

Joseph Camilo, *Student Member, IEEE*, Leslie Collins, *Senior Member, IEEE,* and Jordan Malof, *Member, IEEE,*

*Abstract*—Forward-looking ground-penetrating radar (FLGPR) has recently been investigated as a remote sensing modality for buried target detection (e.g., landmines). In this context, raw FLGPR data is beamformed into images and then computerized algorithms are applied to automatically detect subsurface buried targets. Most existing algorithms are supervised, meaning they are trained to discriminate between labeled target and non-target imagery, usually based on features extracted from the imagery. A large number of features have been proposed for this purpose, however thus far it is unclear which are the most effective. The first goal of this work is to provide a comprehensive comparison of detection performance using existing features on a large collection of FLGPR data. Fusion of the decisions resulting from processing each feature is also considered. The second goal of this work is to investigate two modern *feature learning* approaches from the object recognition literature: the bag-of-visual-words and the Fisher vector for FLGPR processing. The results indicate that the new feature learning approaches outperform existing methods. Results also show that fusion between existing features and new features yields little additional performance improvements.

## I. INTRODUCTION

FORWARD-LOOKING ground-penetrating radar (FLGPR) is a remote sensing modality that has recently been investigated for detecting buried targets (e.g., landmines) [1]–[6]. In this context, FLGPR systems generally consist of an array of radar transmitters and receivers mounted on the front of a vehicle. As the vehicle travels forward along a road or path, radar pulses are emitted towards the ground and the receivers receive energy reflected from the surface and subsurface. The raw data is used in a beamforming process to synthesize 2-dimensional spatial images of the ground (described in Section II.B) [7]–[9]. Pixel intensities in the resulting images can be viewed as a crude measure of the energy reflected from the ground at that location. Figure 1 shows an illustration of the FLGPR detection system considered in this work.

Buried targets can be detected in beamformed FLGPR images because they often reflect the radar energy differently than the surrounding soil and other subsurface clutter (e.g., rocks, roots, etc.) [10]. Examples of beamformed images over target and clutter objects are shown in Figure 2. Although the targets in Figure 2 are easily detectable, many targets exhibit much subtler patterns, and are easily confused with clutter.

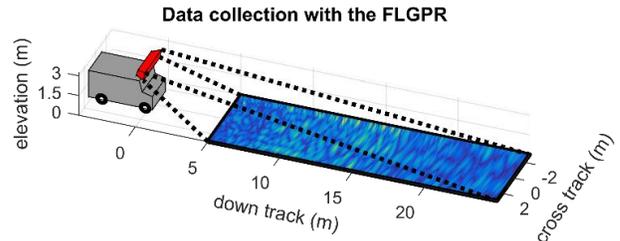

**Data collection with the FLGPR**

Figure 1: A diagram of the FLGPR system. This system inspects the ground in front of the vehicle; the responses from the antenna array are then formed into images for detection. *Cross track* and *down track* labels are used to denote the axes perpendicular and parallel to the vehicle's direction of travel, respectively.

This work considers the application of computerized algorithms to beamformed images in order to automatically detect buried targets. A large body of research has been conducted on this topic [3]–[6], [9], [11]–[15]. Such object detection algorithms usually employ supervised machine learning classifiers (e.g., the support vector machine [16], logistic regression [16]) to distinguish between target and non-target FLGPR imagery. For image recognition tasks such as the one considered here, classifiers operate on image features, statistics, or other measures that are computed based on the images. The performance of a classifier depends strongly on the features it is given, and as a result, a variety of features have been investigated for target detection with FLGPR. Recent examples include SIFT descriptors [15], the 2D FFT of the images [3], local image statistics [15], log-Gabor filtering statistics [4], and raw pixel intensities [5].

Although many features have been investigated, it is still unclear which features are best for detection in FLGPR. The aforementioned existing features were each examined under different conditions: using different data, classifiers, and experimental designs. Additionally, several important recent advances in feature extraction from the computer vision community have yet to be investigated for this problem.

### A. Contributions of this work

To address the above problems, this work makes two primary contributions. First, a comprehensive comparison of detection performance using existing FLGPR features is conducted on a large data set using several supervised classifiers. Second, we

This work was supported by the U.S. Army RDECOM CERDEC Night Vision and Electronic Sensors Directorate, via a Grant Administered by the Army Research Office under Grant W909MY-11-R-0001.

The authors are with the Department of Electrical and Computer Engineering, Duke University, Durham NC 27708

compare the detection performance obtained using existing feature sets to performance attained using two recent successful feature learning methods: the bag-of-visual words (BOV), and the Fisher vector (FV) [17]–[24]. These feature learning approaches have become very popular for image recognition tasks in recent years, and yield excellent performance in a variety of application areas [22], [24], [25]. In addition to evaluating each feature's performance individually, the decision-level fusion of features is also considered.

Experiments were conducted using a large FLGPR dataset consisting of 10 passes over three different test lanes with a vehicle-mounted FLGPR system ($36,000 \ m^2$ of total surface area scanned). The results show that the Fisher vector and BOV feature learning approaches outperform any individual feature set, though that fusing the feature learning decisions with other feature sets yields little additional performance improvement. An analysis of the results provides insight about which image structures in FLGPR data are most indicative of the presence of a target.

The remainder of this paper is organized as follows. Section II describes the FLGPR system and the dataset used for the detection experiments. Section III summarizes the previous features and algorithms used for detection on FLGPR data. Section IV presents a description of the proposed feature learning approaches, and Section V presents the experimental design and results. Section VI provides conclusions and discusses potential future work.

**Example FLGPR radar images**

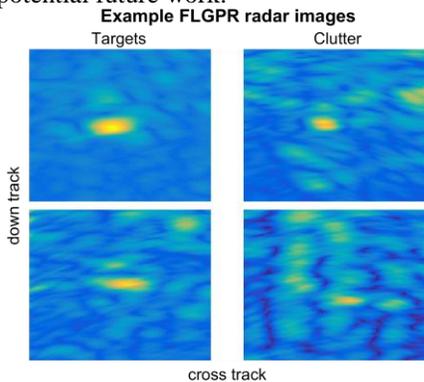

Figure 2: Beamformed images over two different target locations (left column), and two different clutter objects (right column).

## II. THE FLGPR SYSTEM AND DATA

This section first describes how the raw radar data is collected and formed into images (i.e., beamforming), followed by a description of the testing dataset.

### A. The FLGPR radar

The data used in these experiments comes from an FLGPR system that employs a bi-static antenna array, and inspects the ground using a series of stepped frequency pulses [8], [11], [14]. Frequencies are emitted and collected one at a time for each transmit – receive antenna pair. The magnitude and phase change for each emitted frequency is measured and converted, with an inverse Fourier transform, into a corresponding time-series signal. A collection of time-series are then used to synthesize FLGPR images in the beamforming process (Section II.B).

In this FLGPR system, the ground is inspected with L-Band frequencies. Similar to the downward-looking GPR, this frequency range was chosen for its ability to penetrate the ground as well as reflect from target objects [6], [8], [9]. Using a stepped frequency sampling scheme, the L-Band is sampled in 2,702 frequency steps [9].

The system uses three separate polarization schemes, HH, VV, and VH. The first letter in this notation corresponds to the transmitted polarization, and the second corresponds to the polarization measured by the receiving antenna. Polarization is a categorization for the orientation of the electrical component of the electromagnetic signal: the horizontal (H) or vertical (V) orientation. All the vertical antennas are evenly spaced, in one row, across the top of the vehicle. The horizontal antennas are split into three rows, evenly stacked vertically; and in each row antennas are evenly distributed across the vehicle.

### B. Beamforming to create images

Beamforming refers to the process of synthesizing images from time-series returned by the FLGPR transmit-receive antenna pairs. The beamforming process improves the SNR of the radar signals by averaging over the returns of multiple antenna pairs, as well as providing some information about the shape and size of objects [7]. Each beamformed image can be thought of (crudely) as a map of the radar energy reflected from the ground over some spatial extent in front of the vehicle. In this work, we beamform images with similar size and resolution to those in [5], [26]. Figure 3 provides three consecutive frames of beamformed FLGPR data.

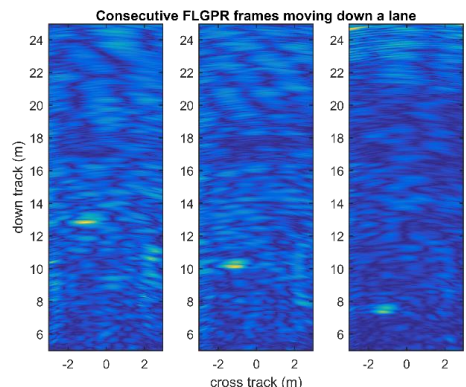

Figure 3: Sequential *frames* in the HH polarization, formed as the sensor moves forward from the left most image to the right most image. Notice the target signature moves towards the bottom of the radar image (closer to the sensor array).

### C. The experimental dataset

Data used in these experiments was collected at a western U.S. Army testing facility. This test site had 245 target (buried threat) encounters in 3 unique test lanes, with a total scanned area of $36,000m^2$. The precise location of each buried target was recorded so that the detection algorithms could be scored. Each buried target is encountered by the FLGPR system multiple times in the data collection, because some lanes are scanned multiple times in both directions. Table 1 includes more details about the lanes and their respective target populations.

Table 1: Details about the data collection used in the experiments. The "total" column is computed using all passes for each lane.

|  | Lane A | Lane B | Lane C | Total |
|---|---|---|---|---|
| **Lane passes** | 4 | 4 | 2 | 10 |
| **Lane area ($m^2$)** | 3,943 | 3,610 | 2,961 | 36,361 |
| **Unique targets** | 28 | 23 | 27 | 245 |
| **Metal** | 9 | 9 | 10 | 90 |
| **Low-metal** | 19 | 14 | 17 | 155 |

## III. BACKGROUND METHODS

This section presents details of the methods used in the experiments presented in Section V. It begins with a description of the FLGPR detection processing pipeline used in this work. This is followed with a more detailed description of the major components in the processing pipeline. Then a brief description is provided for each of the existing features and classifiers employed in the experiments.

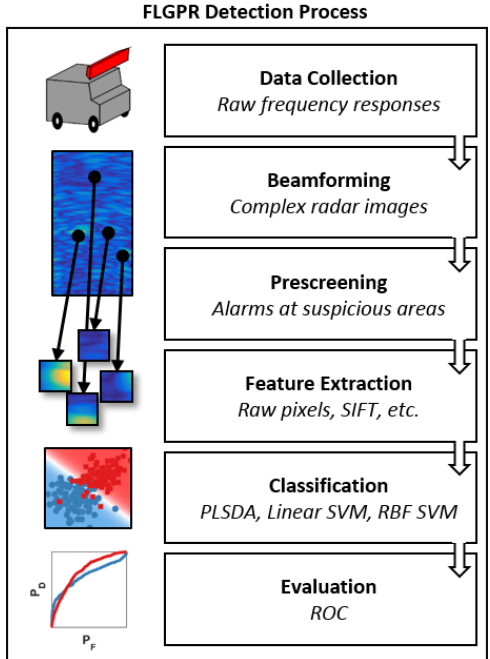

Figure 4: Processing chain for the FLGPR system. Each block represents a major step in the processing pipeline. To the left of each step is an illustration of the neighboring process step.

### A. Overview of detection processing for FLGPR

The detection processing pipeline considered in this work is shown in Figure 4. This pipeline is representative of those employed in many FLGPR studies [3]–[6], [12], [15], [27], [28]. The first step of the pipeline is beamforming, which was described already in Section II.B. Once the FLGPR data is beamformed into radar images, a *prescreener* is run on the beamformed images in order to identify a subset of suspicious locations for further, more sophisticated, processing [29], [30]. The result of prescreening is a list of alarm locations and a decision statistic, or confidence value, indicating how likely each alarm is to correspond to a target.

The prescreening step is followed by feature extraction, where statistics or other measures are extracted from the imagery surrounding each alarm location. These features are provided to the next stage of processing, classification, where a trained machine learning classifier is used to assign a new (and hopefully improved) confidence to the alarm. The output of the classifier stage is a list of alarm locations (the same as the prescreener), but now with the assigned classifier confidence.

The next step of processing is the performance assessment. The score for a particular algorithm (i.e., a combination of features and a classifier) is computed using receiver operating characteristic (ROC) curves. This performance metric is commonly used for buried target detection algorithms in FLGPR [3]–[6], [9], [11]–[15].

### B. The prescreener

The prescreener here is based on the work presented in [5], and is similar to other prescreeners that have been applied to this problem [3]–[6]. It consists of several steps, beginning with the RX algorithm [31]. RX is a constant false alarm rate (CFAR) detector that identifies anomalous data by comparing the statistics of pixels in a foreground window with statistics computed in a background window. This computation is repeated at every pixel location, replacing each pixel with a confidence, resulting in a "confidence image". More precisely, the confidence value at each pixel location in the new confidence image is given by, $\lambda = (\mu_t - \mu_b)^2 / \sigma_b^2$ where $\mu_t$ is the mean of the foreground pixels, $\mu_b$ is the mean of the background pixels, and $\sigma_b^2$ is the variance of the background pixels. Prescreener alarm declarations are made at local maxima locations in the new confidence image. Windows of 40 x 40 pixels and 80 x 80 pixels were used for the foreground and background statistics, respectively. The prescreener is applied to images of the VV polarization, in similar fashion to previous studies [3], [5], [12].

As noted above, due to the multi-look nature of the beamformed images, each area of the ground appears in the beamformed imagery multiple times as the vehicle moves down the lane. This means that the prescreener will often return multiple alarms for a single object (e.g., buried targets, clutter, etc.) seen by the radar. To exploit this multi-look information, the alarms are clustered according to their spatial location (i.e., Universal Transverse Mercator coordinates). For this purpose, a clustering algorithm is used that enforces a limit on cluster radius, in order to maintain the locality of alarms. DP (Dirichlet Process) means is an extension of K-means that can enforce a limit on the cluster radius [32], and is implemented here, with a cluster radius of 1 meter. Each cluster center is retained as an alarm, and the confidence of each alarm is given by the $l_2$-norm of the cluster member confidences [5], [26].

### C. Feature Extraction

All features were extracted from patches beamformed at prescreener alarm locations. A $3m$ x $3m$ patch was beamformed centered over each prescreener alarm, at a down-track distance of 5 meters, and with a spatial resolution of $3cm$/pixel. These specifications are similar to those in previous FLGPR studies [4], [5], [15].

Before features are extracted from an image patch, $X$, each patch is normalized by the local background statistics [4], [5], [15]. A normalized patch, denoted $X'$, is computed by

$$X' = \frac{|X| - \mu_{bg}}{\sigma_{bg}}, \qquad (1)$$

where $\mu_{bg}$ is the mean of the background, and $\sigma_{bg}$ is the standard deviation of the background. The background consists of all of the pixels in the patch, but outside of a 1.5m x 1.5m window centered at the alarm location. All features in this work are extracted on $X'$ unless it is stated otherwise.

### D. Statistical classifiers

In this work we considered three classifiers: a linear support vector machine (SVM) [33], a partial least squares discriminant analysis (PLSDA) classifier [34], and a nonlinear SVM [33]. Both of the SVM classifiers are used because they accompanied one (or more) of the features in the publication where those features were originally introduced. The non-linear SVM has been used with the 2D FFT and log-Gabor statistical feature [4]. This non-linear SVM uses a radial basis function with the default parameters settings C = 1 and $\gamma = \frac{1}{\# \ of \ features}$. The linear SVM is the conventional classifier with the proposed feature learning approaches as well as being previously used with the following features: raw pixels [5], SIFT [15], and LSTAT [15]. The PLSDA classifier is considered because we found that it generally achieves similar, or better, performance on all of the features, as demonstrated by the results in Section V.A, while having much less computationally cost than either SVM.

### E. Detection scoring metrics

The detection algorithms in this work (i.e., the various feature-classifier combinations) are compared using receiver operating characteristic (ROC) curves. ROC curves provide a measurement of the tradeoff between the target detection rate, $P_d$, and the false alarm rate, FAR, as the sensitivity of the classifier is varied. Here $P_d$ refers to the proportion of observed targets that are detected by the algorithm, and FAR refers to the number of false detections that the algorithm returns per square meter of observed lane area.

A related performance metric is also used in these comparisons, called the partial area under the ROC curve (pAUC) [35], [36]. The pAUC is a summary statistic for the ROC curve, and allows us to more succinctly compare many different algorithms. The pAUC measure is the normalized area under the ROC curve from the origin of the x-axis to a specific FAR. For these experiments a pAUC is measured to a FAR of $0.02 \ FA/m^2$, which corresponds to one false alarm every $50 \ m^2$. The pAUC is normalized so that the minimum and maximum attainable values are 0 and 1, respectively. As the area under an ROC curve increases it reflects the ability of that algorithm to detect more targets, within the specified range of FAR values. Both the ROC curve, and the pAUC statistic, have frequently been used to evaluate the performance of buried target detection algorithms for the FLGPR [5], [6], [12], [15], [37].

### F. Previously proposed FLGPR features

In this section we present a brief review of each of the previously proposed FLGPR features that we investigated in this work. Throughout this section, we will use $\psi$ to denote the feature vectors that are delivered to a classifier for each method.

#### 1) Raw pixels

This feature consists of rasterizing the pixels in $X'$, and treating them as a feature vector. More precisely, the raw pixel feature is given by $\psi_{Raw}(X') = \text{vec}(X')$, where the $\text{vec}(\cdot)$ operator refers to the vectorization of a matrix. This type of feature is often used as a simple benchmark in image recognition tasks [17], [20] and has previously been applied to FLGPR [5], [6], [26].

#### 2) Scale invariant feature transform (SIFT)

The SIFT feature aggregates the gradients over regions in an image into a histogram [38]. The first step in computing the SIFT descriptor involves calculating the gradient magnitudes and orientations (of pixel intensities) at each location in the radar image, $X'$. Below, (2) and (3) show the gradient magnitude, $M(i,j)$, and gradient orientation, $\theta(i,j)$, calculations with $i$, and $j$ indexing the pixels of the image.

$$M(i,j) = \sqrt{\begin{array}{l}\left(X'(i+1,j) - X'(i-1,j)\right)^2 + \\ \left(X'(i,j+1) - X'(i,j-1)\right)^2\end{array}} \qquad (2)$$

$$\theta(i,j) = \tan^{-1}\frac{\left(X'(i,j+1) - X'(i,j-1)\right)}{\left(X'(i+1,j) - X'(i-1,j)\right)} \qquad (3)$$

With these gradient calculations, the orientations are then aggregated into 4 by 4 non-overlapping cells. In each aggregation cell, the histogram of orientations is computed. This histogram separates angles into 8 bins between 0 and 360 degrees. The histogram count for each angle bin is computed within each cell and is based on the magnitude of the gradients. There are 16 total cells and 8 angle bins resulting in a 128-dimensional descriptor vector, $\psi_{SIFT}(X')$. The final feature vector is this SIFT descriptor computed over the whole alarm image.

The SIFT descriptor has two roles in this work. First, it is employed in the feature learning approaches described in Section IV.A. We also use it here as a proxy feature for the Histogram of Oriented Gradients (HOG) features due to their similarity. The HOG feature has been previously applied to FLGPR data for buried target detection [5], [15], and therefore it is included here (via SIFT).

#### 3) Local statistics (LSTAT)

Local statistics over an alarm patch are often used in the natural object detection literature [39] and has previously been applied to FLGPR data [15]. To compute an LSTAT feature, the patch is divided into a 3 by 3 grid of non-overlapping regions. The feature vector consists of the mean and variance of the pixel intensities in each of the grid regions. Mathematically this can be expressed by,

$$X' = \begin{bmatrix} x'_1 & x'_2 & x'_3 \\ x'_4 & x'_5 & x'_6 \\ x'_7 & x'_8 & x'_9 \end{bmatrix}, \qquad (4)$$

$$\psi_{LSTAT}(X') = \{[\mathbb{E}\{x'_r\}, Var\{x'_r\}]; \ r = 1, \dots, 9\}. \qquad (5)$$

### 4) 2-Dimensional fast Fourier transform (2D FFT)

This feature consists of the upper right quadrant of the 2D FFT of the FLGPR alarm patch [3], [4], [40]. In contrast to the other features, the 2D FFT is computed on the complex patch, $X'$. Before the 2D FFT, a Hamming window ($H$) is applied to $X'$, and the real component is taken. Equation (6) precisely defines the final feature.

$$\psi_{FFT}(X') = \text{vec}(|FFT_{2D}\{Re(H \circ X')\}|), \qquad (6)$$

where $FFT_{2D}$ refers to the 2D FFT of an image, and $\text{vec}(\cdot)$ refers to the vectorization of a matrix.

### 5) Log-Gabor statistical feature

The log-Gabor filter bank is intended to localize frequency information in an image [41]. These filters were applied to the FLGPR data to potentially improve upon other frequency based features such as the 2D-FFT [4]. The implementation for FLGPR data uses statistics about the log-Gabor filter responses to build the feature vector for a given observation. Here we provide a brief summary of the log-Gabor features, the full details can be found in [4], The log-Gabor filter bank consists of six orientations and six scales, resulting in 36 different filters. The response of an observation, $X'$, to the $i^{\text{th}}$ log-Gabor filter is denoted by $S^i(X')$. Statistics are extracted over 3 by 3 non-overlapping regions for each filtered image. Mathematically we note these grid regions as,

$$S^i(X) = \begin{bmatrix} s^i_1 & s^i_2 & s^i_3 \\ s^i_4 & s^i_5 & s^i_6 \\ s^i_7 & s^i_8 & s^i_9 \end{bmatrix} \qquad (7)$$

Statistics about these regions are then taken,

$$\begin{aligned} &f_{LG}(s^i_r) \\ &= [\mathbb{E}\{s^i_r\}, Var\{s^i_r\}, \text{Kurt}\{s^i_r\}, \text{Skew}\{s^i_r\}, \|s^i_r\|] \end{aligned} \qquad (8)$$

The statistics from each region in the log-Gabor responses are then concatenated to form the final log-Gabor feature vector.

$$\psi_{LG}(X) = \{f_{LG}(s^i_r); \ r = 1, \dots, 9 \ ; i = 1 \dots 36\} \qquad (9)$$

## IV. Applied feature learning approaches

This section presents the two feature learning approaches we investigate in this work: bag-of-visual-words (BOV) and Fisher vector (FV). In contrast to the previously proposed features, these methods automatically infer parameters using training data, and therefore must be trained. Figure 5 provides a high-level overview of the training/testing process, which is similar for both of the two approaches that were investigated.

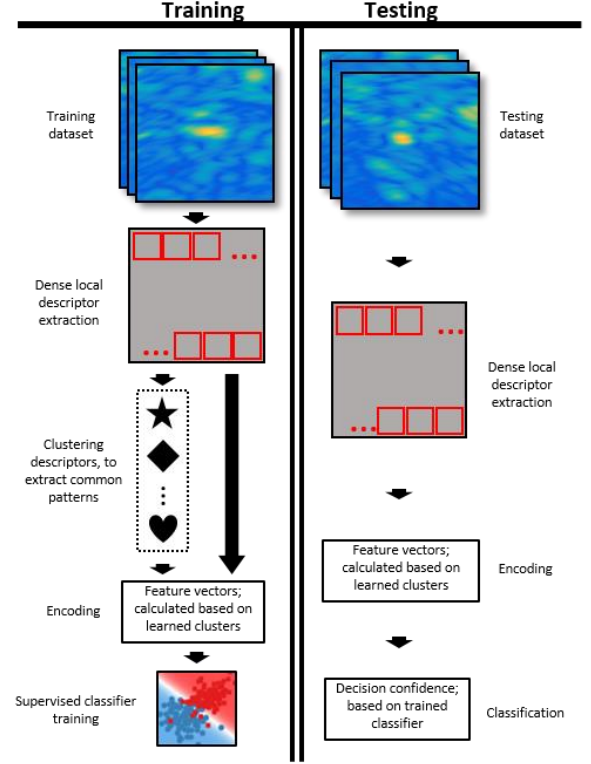

Figure 5: A high-level diagram for the type of feature learning used in this work. First, in both training and testing, each approach requires that descriptors are densely extracted in small overlapping windows over each FLGPR image. During training, these descriptors are clustered to learn a codebook, or dictionary, of common patterns encountered in the data. The dictionary is then used to "encode" each individual FLGPR image, and this encoding acts as the feature vector for that image. This same procedure is repeated during testing to extract a feature vector, except that no clustering is needed.

Details of the steps in Figure 5 are described further in the subsequent subsections. First, we will describe the (two) types of dense descriptors extracted in the images. Next, we describe the clustering and encoding processes for each of the two methods, BOV and FV, respectively. Lastly, we provide some additional implementation details, explaining the specific design choices we made to adapt the BOV and FV approaches to work effectively for target detection on FLGPR data.

### A. Local image descriptors

The feature learning techniques in this work are used with local descriptors that are densely sampled over the alarm patches. In this section, we define an observation (i.e., an FLGPR patch) as a set of local descriptors, $X = \{x_t \in \mathbb{R}^D; \ t = 1, \dots, T\}$. The local descriptors for this work are sampled densely around the image in overlapping sub-patches, as shown in Figure 5. Two popular local descriptors, from the BOV and FV literature, are considered in this work: raw pixels ("Raw"), and SIFT ("SIFT") descriptors [18], [20], [25], [38], [42], [43]. The Raw descriptor is simply a vector containing the raw pixel intensities in each sub-patch. The SIFT descriptor was described in Section III.F.2), and measures local gradient information. Specific details about the implementation of both local image descriptors are provided in Section IV.D.

### B. Bag-of-visual-words (BOV)

Our implementation of BOV is based on Coates and Ng, 2012 [18], but is adapted for target detection in FLGPR data. There are two main components of the BOV implementation: the dictionary creation, and the encoding of observations.

To train the BOV algorithm, local descriptors from all training observations are clustered using Spherical K-means. This yields a set of representative data signals, referred to as a *dictionary*, $\mathcal{D} \in \mathbb{R}^{K \times D}$. Following the suggestion of [18], the descriptors are whitened with zero component analysis (ZCA) before applying spherical K-means. ZCA whitens by projecting the descriptors onto an orthogonal basis. This tends to result in more independent dictionary elements, a desired trait in BOV [18].

The BOV encoding (i.e., feature vector) consists of a similarity measurement of $X$ to the learned clusters, or dictionary elements. The similarity measure is given by

$$\gamma_t(k) = \mathcal{D}_k x_t, \tag{10}$$

where $\mathcal{D}_k$ is the $k^{\text{th}}$ element of the dictionary. The feature vector for an FLGPR image is $K$ dimensional, and the $k^{th}$ feature value consists of the maximum inner product across all descriptors for the given FLGPR image, this is given by

$$\psi_{BOV}(X \,|\, \mathcal{D}) = \left\{ \max_t \{ \gamma_t(k) \} ; k = 1 \dots K \right\}. \tag{11}$$

Notice that the encoding in (12) does not encode any information about the spatial location of descriptors, and therefore the BOV descriptor only encodes *what* is in the image, not *where* it is. Spatial information can be included in the encoding using spatial pooling, discussed in Section IV.D.

### C. Fisher vector

Similar to BOV, the FV is designed to measure the occurrence of learned structures in a single observation. The FV, as implemented here, is based on Sanchez et al., 2013 [25]. Similar to BOV, during training, the FV begins with a clustering operation on the densely extracted local descriptors. Rather than K-means however, FV employs a K-component Gaussian Mixture Model (GMM). Inferring the parameters of the GMM requires finding the means and covariances for each cluster: $\mu_k \in \mathbb{R}^{1 \times D}$, and $\Sigma_k \in \mathbb{R}^{D \times D}$. Here the subscript $k$ refers to the $k^{\text{th}}$ component in the GMM. Following common practice, we constrain the covariance matrix to be diagonal, implying that the elements of the local descriptor are independent. A single component of the GMM is given by

$$u_k \triangleq \mathcal{N}(\mu_k, \Sigma_k), \tag{12}$$

and the GMM is expressed as

$$u_\lambda \triangleq \sum_{k=1}^{K} w_k u_k, \tag{13}$$

where $w_k$ refers to the probability of each cluster (and therefore $\sum_k w_k = 1$), and $\lambda$ denotes the set of parameters learned for the GMM. In other words, $\lambda = \{ w_k, \mu_k, \Sigma_k; \; k = 1, \dots, K \}$.

Once the codebook (i.e., trained GMM) is available from training, it can be used to encode (i.e., compute a feature vector) for a new observation. The encoding (roughly) consists of computing first and second order differences between each of the $T$ descriptors, $x_t$, and the cluster centers. This computation is given by equations (14) and (15) below:

$$\mathcal{G}_{\mu_k}^X = \frac{1}{\sqrt{w_k}} \sum_{t=1}^{T} \gamma_t(k) \left( \frac{x_t - \mu_k}{\sigma_k} \right), \tag{14}$$

$$\mathcal{G}_{\sigma_k}^X = \frac{1}{\sqrt{w_k}} \sum_{t=1}^{T} \gamma_t(k) \frac{1}{\sqrt{2}} \left[ \frac{(x_t - \mu_k)^2}{\sigma_k^2} - 1 \right]. \tag{15}$$

The function $\gamma_t(k)$ in the preceding two equations is given by

$$\gamma_t(k) = \frac{w_k u_k(x_t)}{\sum_{j=1}^{K} w_j u_j(x_t)}. \tag{16}$$

In (15) and (16), and given a particular value of $k$, the terms $\mathcal{G}_{\mu_k}^X$ and $\mathcal{G}_{\sigma_k}^X$ are each vectors of length $D$, where $D$ is the dimensionality of the input descriptor (e.g., Raw or SIFT). The final FV feature is given by

$$\psi_{FV}(X \,|\, u_\lambda) = \{ \mathcal{G}_{\mu_k}^X, \mathcal{G}_{\sigma_k}^X, k = 1..K \}, \tag{17}$$

and its dimensionality is 2DK.

It is important to notice that each encoding $\mathcal{G}_{\mu_k}^X$ and $\mathcal{G}_{\sigma_k}^X$ consist of a weighted sum, or "pooling", of contributions from individual descriptor differences. This implies that the spatial location of each descriptor is lost in the computation of the FV feature. Consequentially, and similar to the BOV encoding, the FV feature only encodes information about *what* is in each FLPGR image, but not *where* it exists in the image. This is addressed through a technique called spatial pooling, which we describe next in Section IV.D.

### D. Additional implementation details

This section presents some additional details about our implementation of BOV and FV in order to adapt them to target detection in the FLGPR imagery. We first describe the details of the local descriptor extraction. Second, we describe the number of clusters, or components, used for these feature learning methods. Lastly, we describe our application of spatial pooling to enhance the performance of both the BOV and FV features.

In this work the BOV and FV methods are each applied using two types of local descriptors. First, raw radar image intensities of 11 x 11 pixel regions were densely sampled over the FLGPR images, with a stride of 7 pixels. These sizes were chosen because they performed the best for both BOV and FV. This first descriptor will be referred to as BOV (Raw) and FV (Raw). The other local descriptor tested was SIFT, which was extracted every 8 pixels, and over 8 x 8 pixel regions, for each background normalized magnitude alarm patch. This size and stride were also chosen to maximize performance in cross-validation. Once again, the same settings yielded the best results for both the FV the BOV features. The SIFT-based BOV and FV features will be denoted as BOV (SIFT) and FV (SIFT), respectively.

For each encoding scheme 30 component clustering was used, and this was done for both types of descriptors: SIFT and Raw. We found that changing the number of components for both BOV and FV in the 10 – 50 component range yielded very similar performance.

Spatial pooling is also applied to BOV and FV for both types of local descriptors. As described in Sections IV.B and IV.C, the BOV and FV features primarily encode *what* is in the image, but not *where* it is. This is a well-known limitation of the BOV and FV approaches [20] and, similar to other recognition tasks, we discovered that spatial information is important for identifying buried threats in FLGPR data. As a result, we adopted a technique called "spatial pooling" [44], which can be used to augment the feature encodings with spatial information. Figure 6 illustrates the concept of spatial pooling, as well as the way we applied it in this work. We found that spatial pooling using a 2 by 2 non-overlapping grid (as shown in Figure 6) resulted in substantial performance improvements, for both feature learning methods (BOV and FV), and for both local descriptors (raw and SIFT).

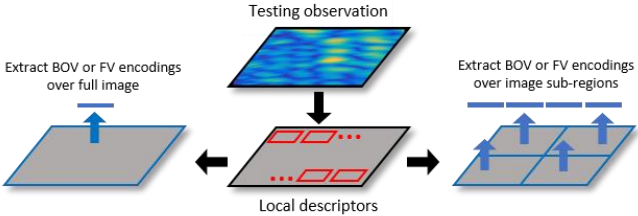

Figure 6: Illustration of different spatial pooling techniques for BOV and FV. In all cases, local descriptors are densely extracted over the FLGPR image. In the original pooling scheme (left), one encoding is computed over the entire image. The encoding discards information about where objects exist in the image. In order to encode this information, we applied spatial pooling in which BOV and FV encodings can be computed on different sub-regions of the image. The resulting encodings (e.g., four in the illustration) are concatenated together to form one long feature vector, where each of the four segments encodes information about the image content in its respective spatial region. In this work we applied a 2 by 2 non-overlapping pooling scheme (as shown), which substantially improved performance.

## V. EXPERIMENTAL DESIGN AND RESULTS

In this section we present the experimental results. We begin by reporting results (in terms of pAUC) for each combination of polarization, feature set, and classification model. After this, we present results where we used a greedy feature selection approach to select a good subset of feature sets for use in decision fusion. Finally, we present an analysis of the results, providing additional insight about the features.

All of the subsequent experimental results were conducted using patches extracted at alarm locations declared by the prescreener described in Section III.B. This prescreener yielded 15,750 alarms (i.e., patches) over the entire dataset. The classification algorithms were tested using a lane-based cross-validation procedure. There were three distinct lanes in our dataset and thus, three-fold cross-validation was used.

### A. The performance of individual feature sets

In this section we present performance results, in terms of pAUC, for each possible combination of (i) radar polarization, (ii) feature set, and (iii) a classifier (e.g., 2D FFT feature of an HH image with the RBF SVM classifier). We refer to such 3-tuples as "algorithms", for ease of discussion. For each algorithm we report an average pAUC, and a 95% confidence interval based on bootstrap aggregation [45]–[48]. Specifically, for each training fold, we created ten different datasets by taking bootstrap samples of the training data (i.e., sampling the original training data, with replacement, until a new equally large dataset is created). A classifier is trained on each of the ten datasets, and then applied to the test set to obtain predictions on the same test set. These ten trials permit us to measure the mean and variance of the performance for each algorithm, which help indicate the consistency and robustness of the algorithm performance. The results of these experiments are presented in three separate figures, where each figure corresponds to a polarization: HH (Figure 7), VV (Figure 8), and VH (Figure 9).

The results reveal a clear trend in the detection performance on each polarity. The HH polarity yields the best mean detection performance across all combinations of features and classifiers, without any exceptions. This is followed by the VV polarity which, in turn, always outperforms the VH polarity.

The results also indicate that there are general differences in the performance of the classifiers. Among the linear classifiers, PLSDA almost always outperforms the linear SVM. The only exceptions to this occur for the SIFT features, and this is mitigated by the overall poor performance of SIFT features. We believe PLSDA produces superior performance due to its ability to deal with collinearity (i.e., redundancy) in the features [34]. Many of the feature sets investigated here are very high dimensional, and occasionally the dimensionality is much greater than the number of observations (e.g., the FV features). This high dimensionality tends to increase the redundancy of the features, likely making PLSDA a more suitable classifier.

In contrast to the linear SVM, the (non-linear) RBF SVM performs similarly to PLSDA. The RBF SVM is nonlinear, and therefore it can model more complex relationships between the features than PLSDA. The performance similarity of the RBF SVM and PLSDA suggests that the greater complexity of the RBF SVM yields few benefits. This further suggests that the patterns in this data are relatively simple. We provide further qualitative support for this assertion in Section V.D. Although PLSDA and the RBF SVM perform similarly, the RBF SVM is much more computationally expensive (i.e., slower) than PLSDA, during both training and testing, and so we generally favor PLSDA.

For the remaining discussion we will only consider results with the RBF SVM and PLSDA, since they both (almost) always outperform the linear SVM. When comparing individual features, the results reveal several trends. First, many algorithms are outperformed by the raw pixel features. Secondly, overall the best performing features *tend* to be the FV-based features, raw, and LSTAT. The best overall mean performance is achieved with FV (SIFT) for the RBF SVM.

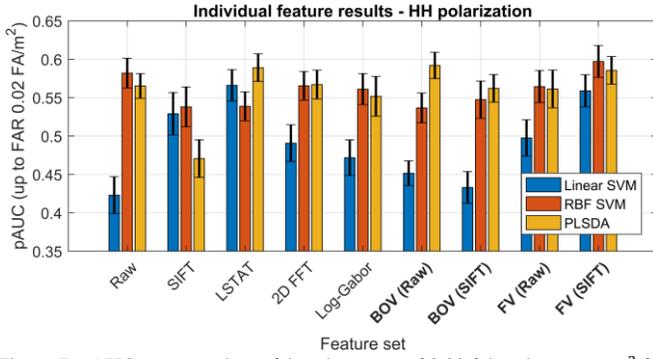

Figure 7: pAUC, computed to a false alarm rate of 0.02 false alarms per $m^2$ for each feature set on the HH polarization using each considered classifier. Feature learning approaches are bolded. The mean pAUC and 95% confidence intervals for 10 bootstrap trials are reported.

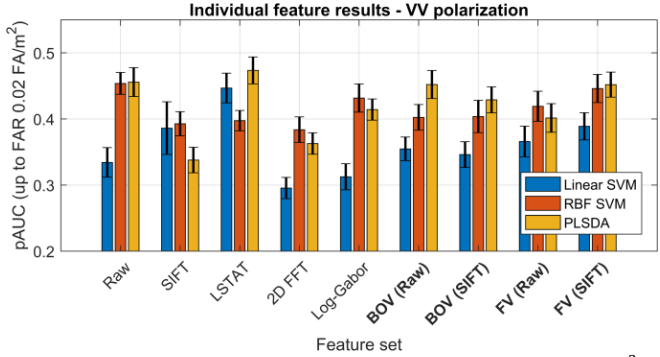

Figure 8: pAUC, computed to a false alarm rate of 0.02 false alarms per $m^2$ for each feature set on the VV polarization using each considered classifier. Feature learning approaches are bolded. The mean pAUC and 95% confidence intervals for 10 bootstrap trials are reported.

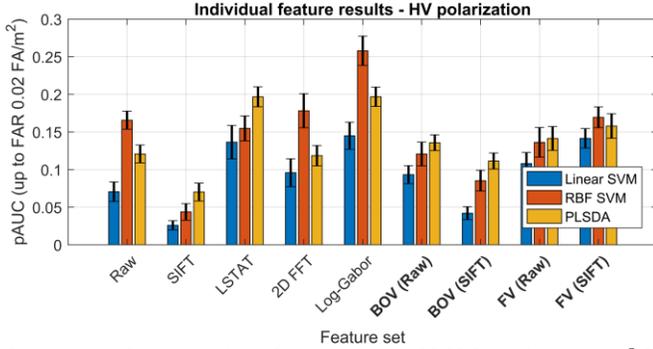

Figure 9: pAUC, computed to a false alarm rate of 0.02 false alarms per $m^2$ for each feature set on the HH polarization using each considered classifier. Feature learning approaches are bolded. The mean pAUC and 95% confidence intervals for 10 bootstrap trials are reported.

### B. Decision-level fusion

In decision fusion we aim to combine the predictions (i.e., confidence values output by a classifier) from the different algorithms in Section V.A in order to further improve detection performance [49]. Decision fusion was also investigated recently for FLGPR in [50], and we build on this work by using a larger collection of data and a greater variety of image feature sets.

In order to fuse the algorithm predictions, we treat them as features that are input into a second PLSDA classifier, which is then trained to make a final prediction. For simplicity, in these experiments we only consider fusing the PLSDA classifier predictions, but we consider all polarities and feature sets. It is unlikely that all of these feature sets are useful for fusion, and so we attempt to select a good subset of them for fusion. To select this subset we used the sequential forward search algorithm (SFS) [51]. This algorithm begins with the predictors (i.e., the PLSDA predictions) for the single best feature set, and then adds a new predictor one at a time, based on which one increases the performance the most. The SFS is only allowed to select new predictors based upon the training data in each fold, in order to avoid positive performance bias. Therefore, in order to obtain performance measures for each candidate fusion model, we perform a random five-fold cross-validation using only the training data. This yields a pAUC that SFS can use to build the final fusion model.

In our experimental design we control the total number of predictors that SFS is allowed to select for fusion, denoted $N_f$. Figure 10 presents the results of our feature fusion experiment as we vary $N_f$. Each point in Figure 10 represents the mean of the pAUCs, and the error bars report the 95% confidence interval [48]. This interval is computed after repeating each experiment ten times to account for the randomness introduced by the 5-fold cross-validation within each training fold. The results show a sharp performance increase at $N_f = 2$ then a steady rise until $N_f \cong 20$.

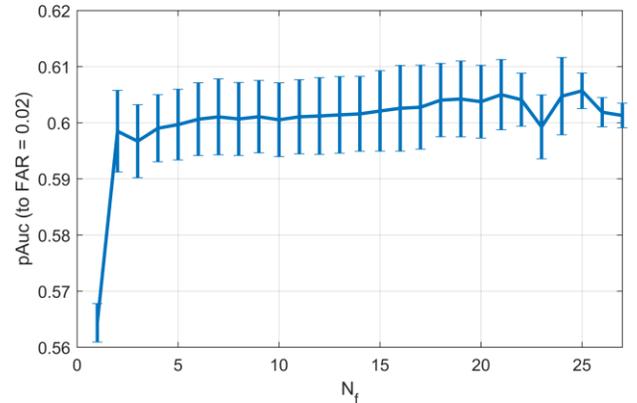

Figure 10: pAUC for decision fusion over a varying number of features for the sequential forward search. Each point if the mean performance for that number of feature and the error bars are the 95% confidence intervals over multiple runs of the same experiment; for each of the number of features we ran the experiment 10 times. Notice the leveling of performance after allowing two features to be selected; adding more features past this only slightly increases overall performance.

To estimate overall performance for the SFS algorithm we conducted an experiment using an auto-stopping criteria to select $N_f$. The algorithm would stop increasing $N_f$ when the pAUC within that fold started to decrease. The pAUC is determined from the random five-fold cross-validation performed within each training set, thus this experiment was repeated 10 times to better estimate the performance. In Figure 11 the vertically averaged ROC [52] and 95% confidence intervals are reported for the SFS auto-stopping algorithm and the FV (SIFT) with the RBF SVM, along with the performance of the prescreener. The decision fusion and best individual feature set yield very similar performance at any given point on the ROC curve.

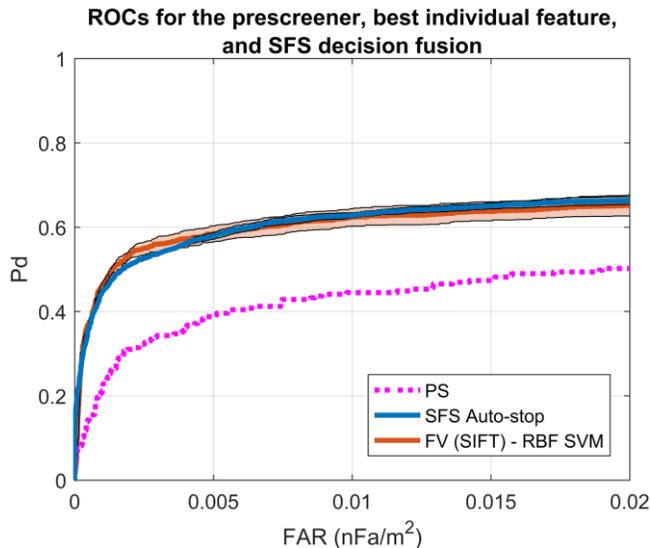

**ROCs for the prescreener, best individual feature, and SFS decision fusion**

Figure 11: ROC curve (probability of detection vs. false alarm rate) for prescreener (PS), the single best feature (FV Raw – RBF SVM), and the sequential forward search (SFS) decision level fusion results.

### C. Learned filters

Feature learning algorithms can be analyzed by visualizing the resulting filters from the clustering, and this is done here in Figure 12 for the BOV (Raw) algorithm on the HH polarization. We analyze BOV (Raw) because (i) it was nearly the best performing algorithm and (ii) it uses the raw pixel descriptor which results in more interpretable visualizations. During feature extraction the resulting BOV encodings are based on the similarity between each observation's descriptors and these cluster centers. We can surmise that these cluster centers are useful in discrimination because of the relatively good BOV (Raw) classification performance; that is to say, the space derived with these cluster centers produces a fairly discriminative representation. Notice the "blob-like" learned cluster center, and how many of the clusters are shifted versions of the same shape. This, along with knowing that performance did not increase when introducing more available clusters implies that there are not many cues beyond concentrations of high energy that indicate the presence of a target in FLGPR data.

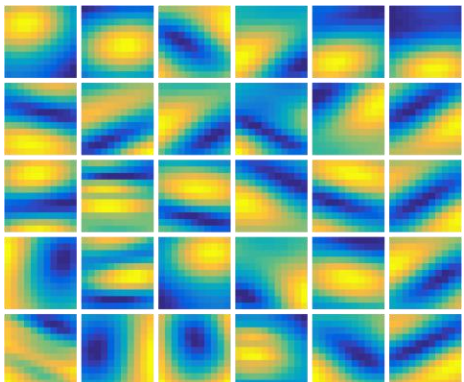

Figure 12: Learned K-means cluster centers for the BOV on raw pixel local descriptors in the HH polarization. With this visualization, it becomes evident that the learned filters are measuring the different "blob-like" characteristics of the data.

### D. What characteristics of FLGPR images indicate the presence of a target?

For the many computer vision algorithms, including feature learning, it can often be difficult to understand which components of the observation images are useful in classification. In an effort to analyze this we developed a "confidence map" visualization based on the feature learning encodings and trained classifier. The subsequent visualizations show the magnitude image and a BOV (Raw) "confidence map" of four target examples over a range of classification confidences. Again, BOV (Raw) on the HH polarization is used here due to its overall good classification performance and the visual interpretability of using raw pixel descriptors.

The "confidence maps" illustrate where, spatially, the BOV (Raw) features indicated target-like characteristics in the image. Figure 13 shows the process for obtaining this visualization. As the process shows, the dense descriptors (i.e., raw patches) in the image are encoded using a sliding window, where four descriptors are encoded at a time. This corresponds to spatial pooling of the BOV encodings (see Section IV.D) over very small image regions. Each encoding is then classified (i.e., assigned a confidence) using a trained BOV(Raw) PLSDA classifier. This process results in a map of confidence values across the image, indicating how much the local descriptors indicate the presence of a target.

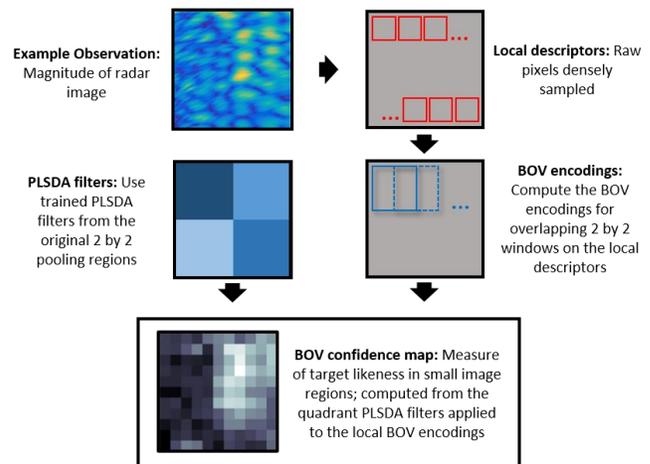

Figure 13: Description of how the BOV (Raw) confidence map is computed for each example observation. Using the learned quadrant PLSDA filters and BOV encodings for 2 by 2 overlapping windows of the local descriptor, a measure of target likeness is calculated for local regions around the observation. In order to encode the extracted descriptors, we used the same K-means clusters that were trained as described in Section V.A. Obtaining an appropriate classifier for the local BOV encodings was more difficult. For this purpose, we used the PLSDA classifier that was trained as in Section V.A, however, that classifier was trained using 2 by 2 non-overlapping pooling. Recall that this pooling scheme yielded one BOV encoding for each quadrant of the image. If the total BOV (Raw) dimensionality is given by $D$, then the PLSDA classifier consisted of $4D$ weights: one set of $D$ weights each of the four quadrants. Therefore, for our discriminability map, if a local BOV encoding was located in the top right quadrant, then the $D$ PLSDA filters weights from that quadrant were applied to the encoding in order to assign a confidence. In the confidence maps, the brighter areas correspond to very target like locations.

For each example in Figure 14, the magnitude images, BOV (Raw) confidence maps, and confidence percentile are given. The confidence percentile denotes the confidence a target was given by the classifier but normalized in relation to the other observations to fit between 0% and 100% (100% confidence percentile being the highest confidence observation). The visualizations in Figure 14 reveal several interesting characteristics about the FLGPR images. First, it appears that the BOV (Raw) feature is largely cuing on high energy "blobs", of varying shape and size, to identify the presence of a target. As expected with the spatial invariance introduced by "bagging" in BOV, the algorithm is able to assign a high confidence to an off-center target response, like that for the (95.6% confidence percentile example. In the last column example (1.3% confidence percentile) there are high confidence locations around a target response but not enough to outweigh the large amount of background present in the image. The example with a 14.0% confidence percentile illustrates a very weak target response, and while some of the response appears target like it is not strong enough for the BOV (Raw) algorithm to classify well.

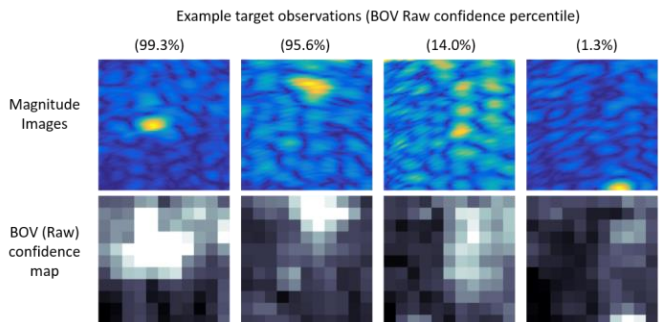

Figure 14: Four target examples with various BOV (Raw) confidence percentiles. Along with magnitude image patch for each, a BOV (Raw) confidence map is included to show where in the image the algorithm found target-like characteristics.

## VI. CONCLUSIONS

In this paper, two contributions to the FLGPR research were presented: a comprehensive comparison of existing features' performance on a large consistent data collection, and the application of feature learning to the FLGPR radar images using the BOV and FV methods. Through these experiments a number of observations were made about the FLGPR. First, that the best performance balanced with computational complexity for any polarization and individual feature set occurs for the BOV with Raw pixel descriptors on the HH polarization. Feature learning in general performed well, but did not outperform all previous feature sets for all polarizations. Analysis of the classification results showed that amorphous "blob-like" structures were the strongest cue for the presence of a target. Second, by fusing the feature decision confidences together, little improvement in classification is achieved. A summary of these conclusions are:

- Using many classifiers and features, the HH polarization imagery consistently yields substantially better performance than imagery based on the VV and VH polarizations.
- Feature learning (Fisher Vectors with the SIFT descriptor, and BOV with Raw pixels) generally yields better performance compared to existing features.
- A linear classifier, PLSDA, which was previously unused in the FLGPR context, consistently yields better performance than a linear SVM. It consistently achieves comparable performance to a nonlinear

SVM, but with much less training time and model complexity.

- Decision fusion across polarities and features resulted in only slightly improved detection performance.

Future work in this area can advance in several directions. First, while we found that it is not very beneficial to combine feature sets at the decision level, this does not necessarily portend that combining feature sets before classification (i.e., feature level fusion) will not lead to greater performance improvements. This may also reveal that there is benefit to fusing different polarities, even though we did not see benefits here when fusing at the decision level.

## ACKNOWLEDGMENTS

This work was supported by the U.S. Army RDECOM CERDEC Night Vision and Electronic Sensors Directorate, via a Grant Administered by the Army Research Office under Grant W909MY-11-R-0001.